\documentclass{article} 
\usepackage{conference,times}

\usepackage{amsmath,amsfonts,bm}

\def\eqref#1{equation~\ref{#1}}
\def\1{\bm{1}}

\DeclareMathAlphabet{\mathsfit}{\encodingdefault}{\sfdefault}{m}{sl}
\SetMathAlphabet{\mathsfit}{bold}{\encodingdefault}{\sfdefault}{bx}{n}

\usepackage{hyperref}
\usepackage{url}

\usepackage{amsmath,amsfonts}
\usepackage{algorithm}
\usepackage{array}
\usepackage{graphicx}

\newcommand{\eg}{\textit{e}.\textit{g}.}
\newcommand{\ie}{\textit{i}.\textit{e}.}

\usepackage{algpseudocode}
\usepackage{amssymb}
\usepackage{float}
\usepackage{booktabs}
\usepackage{wrapfig}
\usepackage{microtype}
\usepackage{multirow}
\usepackage{makecell}
\usepackage{enumitem}
\usepackage{fontawesome5}
\usepackage[percent]{overpic}
\usepackage{xcolor}
\usepackage{tikz}
\usepackage[table]{xcolor}

\usepackage[normalem]{ulem}                      

\title{Uncertainty-Aware Consistency Distillation for Few-Step Video Generation}

\iclrfinalcopy

\author{Lingyu~Liu \\
School of Software \\
Xi'an Jiaotong University\\
Xi'an, China.
\texttt{liulingyu@stu.xjtu.edu.cn} \\
\And
Yaxiong Wang \thanks{Corresponding author.} \\
School of Computer and Information Science \\
Hefei University of Technology \\
Hefei, China. \\
\texttt{wangyx15@stu.xjtu.edu.cn} \\
\AND
Li Zhu \\
School of Software \\
Xi'an Jiaotong University\\
Xi'an, China.\\
\texttt{zhuli@xjtu.edu.cn}
\And
Zhedong Zheng\thanks{Corresponding author.} \\
Faculty of Science and Technology, and Institute of Collaborative Innovation \\
University of Macau\\
Macau, China.\\
\texttt{zhedongzheng@um.edu.mo} }

\begin{document}

\maketitle

\begin{abstract}
We study few-step video generation, \ie, distilling a multi-step video generator, which typically requires tens of sampling steps, incurring substantial latency and compute, into a few-step student. Consistency distillation is a common recipe, in which a multi-step teacher provides the consistency targets for a few-step student. However, these teacher-guided targets are not equally trustworthy, and the content is harder to learn where it varies rapidly over time, \eg, moving foliage shadows or flowing water. 
We observe that supervision reliability follows the local difficulty of the content rather than semantic complexity: regions that change little yield consistent endpoint predictions, whereas regions with large temporal variation produce larger discrepancies that coincide with the largest perceptual errors. Motivated by this observation, we propose \textbf{Uncertainty-Aware Consistency Distillation} (UACD), which reweights consistency supervision at each spatiotemporal region using a local, parameter-free uncertainty estimate. Specifically, we construct two independently perturbed teacher-guided consistency paths, whose student endpoint predictions provide a consensus target; the discrepancy between the student's direct prediction and this target is the uncertainty proxy. We then relax the consistency penalty on high-uncertainty regions through an exponential weight, while keeping the full penalty elsewhere, since the student cannot be expected to match targets that are hard to learn. To preserve perceptual quality under aggressive step reduction, we integrate feature-space adversarial training with semantic alignment. With parameter-efficient LoRA adaptation of the 50-step Wan model, our method achieves state-of-the-art 4-step generation on VBench 2.0 (0.556 mean score) and is preferred over competing methods in a user study\footnote{\tiny Anonymous Github: \url{https://uacd.github.io/UACD/}}.
\end{abstract}

\section{Introduction}
Recent advances in video diffusion models, particularly those based on Flow Matching~\citep{song2020denoising,hunyuanvideo,sora,wan}, can now synthesize high-fidelity and temporally coherent videos from text or image prompts. However, these models typically rely on iterative numerical solvers like Euler or Runge-Kutta methods to traverse the probability flow ordinary differential equation (PF-ODE), which requires tens to hundreds of neural network evaluations per video. This cost translates into high inference latency and limits deployment in real-time applications such as interactive media and game development. 

Step distillation~\citep{meng2023distillation,salimans2022progressive} has emerged as an effective way to cut inference cost while preserving generation quality. Among existing approaches, consistency distillation~\citep{song2023consistency} is particularly effective: the teacher first takes a small Euler step from a noisy state, after which the student predicts the clean endpoint from the resulting intermediate state to form the distillation target. The student is then trained to produce the same endpoint directly from the original noisy state. While consistency distillation works well for image synthesis, applying it to high-dimensional video generation remains challenging: the reliability of the consistency target varies across spatiotemporal regions, and the content itself is harder to learn where it changes rapidly between frames.

\begin{figure*}[t]
 \centering
  %\vspace{-.1in}
 \includegraphics[width=\linewidth]{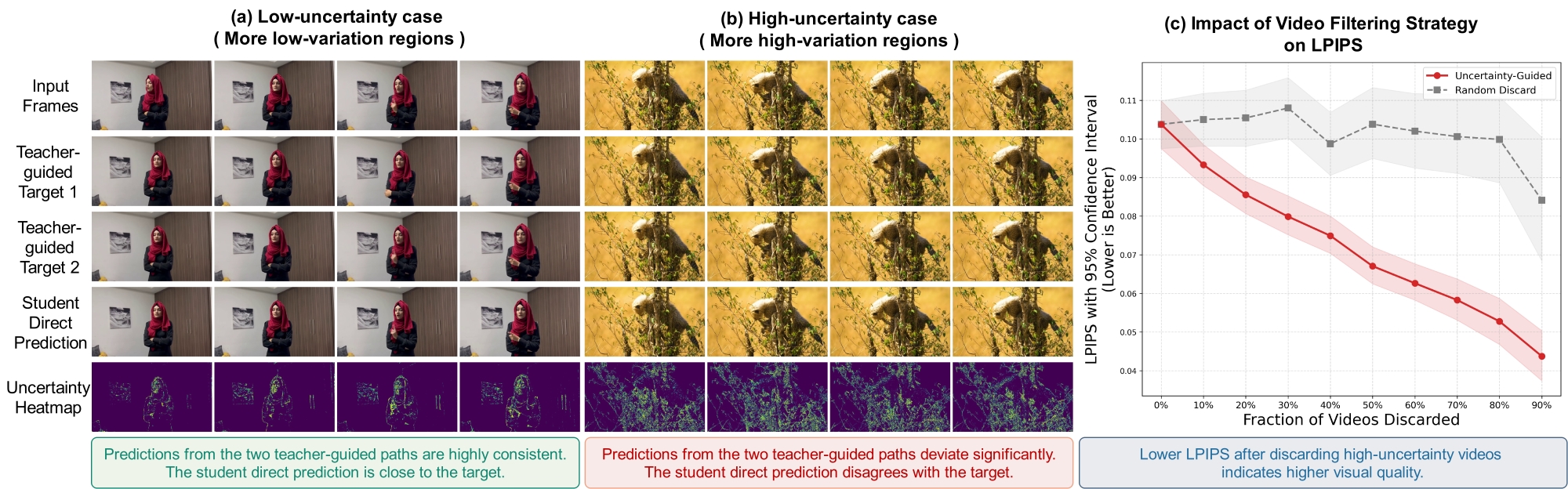}
 \vspace{-.3in}
\caption{Uncertainty-aware supervision according to local consistency reliability. \textbf{(a)} In low-variation regions, two independently constructed teacher-guided consistency paths yield similar student endpoint predictions, producing a small student--consensus discrepancy and thus low uncertainty $U$ (dark heatmap). \textbf{(b)} In high-variation regions, where the content is hard to learn (\eg, moving foliage shadows or flowing water), the two paths produce more divergent endpoint predictions, resulting in a larger discrepancy between the student's direct prediction and the consensus target, and thus higher $U$ (bright heatmap). \textbf{(c)} Risk-coverage analysis showing that uncertainty-guided filtering consistently reduces LPIPS (lower is better) as high-uncertainty videos are removed, while random discarding has little effect. Our loss relaxes the consistency penalty where uncertainty is high and keeps the full penalty elsewhere.}
 \label{fig:teaser}
 \vspace{-.2in}
\end{figure*}

Inspired by recent advances in uncertainty learning~\citep{zhang2025ctrl,wang2026rigi}, we observe that \textbf{not all spatiotemporal regions are equal} in terms of supervision reliability and that the local reliability is tied to how hard the content is to learn. As illustrated in Figure~\ref{fig:teaser}, panels (a) and (b) present the ground-truth videos, two independently perturbed teacher-guided targets, the student's direct prediction, and the resulting uncertainty. Panels (a) and (b) correspond to a low-variation case and a high-variation case, respectively. In the low-variation case, the two teacher-guided predictions and the direct prediction remain consistent, resulting in a stable consensus target and low uncertainty. In contrast, for the high-variation case, these predictions exhibit larger discrepancies, leading to a less reliable consensus target and higher uncertainty. Figure~\ref{fig:teaser}(c) presents a risk-coverage analysis on 300 generated videos, where videos with the highest mean uncertainty $\bar U$ are progressively removed and LPIPS~\citep{zhang2018unreasonable} is measured at each coverage level. Compared with random discarding, uncertainty-guided filtering reduces the average LPIPS of the remaining videos, indicating that $\bar U$ tracks perceptual error and can identify, without access to ground-truth frames, the videos on which distillation is least reliable. These are the videos whose content varies most over time, which suggests that temporal variation, rather than semantic complexity, governs the reliability of consistency supervision. However, conventional consistency distillation applies the same penalty to every spatiotemporal region, regardless of how reliable its target is. Regions that are hard to learn therefore contribute as much gradient signal as regions that are easy to learn, forcing the student to match targets that are themselves unreliable and degrading generation fidelity.

To address this issue, we propose \textbf{Uncertainty-Aware Consistency Distillation}, a few-step video generation framework that adapts the consistency supervision according to fine-grained spatiotemporal uncertainty. Instead of relying on a single teacher-guided perturbation path, we introduce a dual-path consistency construction that enables parameter-free estimation of local consistency uncertainty. Specifically, we construct two independently perturbed noisy states from the same clean latent using two independent noise samples. The teacher performs a one-step Euler update from each perturbed state, producing two distinct intermediate states. The student then predicts the final endpoint from each intermediate state in a single forward pass, yielding two student endpoint predictions whose average forms a consensus target. In parallel, one of the two perturbed states is randomly selected as the input to the direct student prediction. The discrepancy between the student's direct endpoint prediction and the consensus target provides a simple, parameter-free proxy $U$ for local consistency uncertainty, which we then normalize to $\tilde{U}$ for adaptive supervision weighting. This proxy indicates the reliability of the resulting consistency supervision. We then use this signal to formulate an uncertainty-aware consistency loss. With an exponential weight $w=e^{-\lambda\tilde{U}}$ on the distillation objective, where $\lambda$ controls the weighting strength, the consistency penalty is relaxed on regions that are hard to learn, while regions with near-zero uncertainty keep the full supervision signal. Requiring the student to match targets that are themselves hard to learn is unnecessary, and enforcing it uniformly only injects unstable gradients into training. To reduce perceptual degradation under aggressive step reduction, we also employ feature-space adversarial refinement with semantic alignment to preserve high-frequency details and semantic fidelity. The entire framework is trained with parameter-efficient LoRA adapters~\citep{lora}, making it directly applicable to billion-scale video diffusion transformers. We apply uncertainty-aware consistency distillation to the Wan video generation model and reduce inference to four steps while maintaining high visual fidelity. On the VBench 2.0 benchmark, our model achieves superior performance compared to existing few-step distillation methods. User studies further indicate a strong preference for our generated videos. Our main contributions are summarized as follows:
\begin{itemize}[leftmargin=0pt,itemsep=0.5pt]
    \item We identify that the reliability of consistency supervision follows the difficulty of the content, \ie, the magnitude of its temporal variation, rather than semantic complexity, motivating fine-grained, region-wise uncertainty modeling for few-step video distillation.
    \item We introduce a parameter-free local consistency uncertainty proxy based on direct and consensus student predictions, and use this proxy to adapt the supervision strength at each spatiotemporal region according to its local reliability, complemented by adversarial refinement for perceptual quality.
    \item We achieve state-of-the-art 4-step video generation with the Wan backbone on comprehensive VBench 2.0 evaluations (0.556 Mean Score) and are preferred over prior consistency distillation methods in a user study, while maintaining training efficiency through LoRA adaptation.
\end{itemize}

\section{Related Work}
\noindent\textbf{Diffusion Model Distillation.}
Diffusion distillation~\citep{mansourian2025comprehensive,luhman2021knowledge,zheng2023fast} reduces the inference cost of diffusion models~\citep{blattmann2023align,ho2020denoising,ho2022video} through trajectory-preserving~\citep{luo2024latent,ding2025dollar,gu2023boot,wang2024phased,salimans2022progressive,frans2025one} or distribution-matching~\citep{sauer2024adversarial,wang2023diffusiongan,lu2025adversarial,Chen_2026_CVPR,yin2024one} objectives. Video-specific methods include DCM~\citep{dcm}, DOLLAR~\citep{ding2025dollar}, and adversarial self-distillation~\citep{yang2026towards}. Our work complements these approaches with uncertainty-aware consistency distillation that adaptively reweights supervision according to local prediction reliability.

\noindent\textbf{Uncertainty Learning.}
Uncertainty quantification is widely used to improve model reliability and robustness~\citep{kendall2017uncertainties,raghu2019direct,nandy2020towards,lee2020gradients}, with applications in 3D reconstruction~\citep{sabour2023robustnerf}, image classification~\citep{litrico2023guiding}, and image retrieval~\citep{zhang2022implicit,dou2022reliability,chen2024composed,wang2026rigi}. Recent works also exploit uncertainty for reward modeling~\citep{zheng2021rectifying,zhang2021uncertainty,zhang2025ctrl,zhang2024vl} and active view selection~\citep{her2023uncertainty,jiang2024fisherrf}. To our knowledge, uncertainty has not been directly incorporated into the consistency distillation objective for video generation. We address this gap by deriving a local uncertainty proxy from the discrepancy between the student's direct prediction and a dual-path consensus target.

\section{Methodology}

\begin{figure*}[t]
 \centering
 %\vspace{-.1in}
 \includegraphics[width=\linewidth]{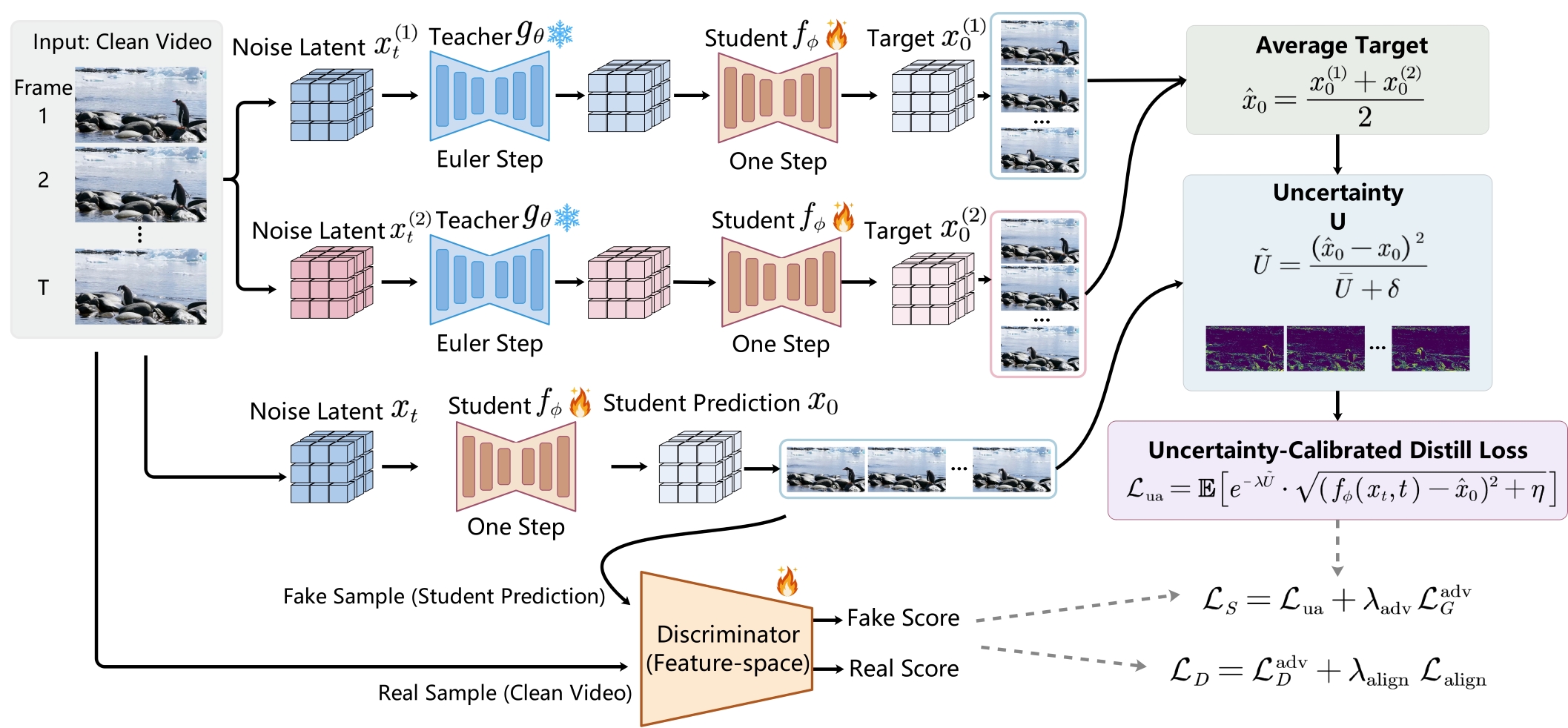}
 \vspace{-.35in}
 \caption{Overview of Uncertainty-Aware Consistency Distillation. Two independently perturbed paths are each advanced by one Euler step using the frozen teacher, producing two teacher-guided intermediate states. The student then predicts the endpoint $x_0^{(1)}$ and $x_0^{(2)}$ from each intermediate state, and their average forms a detached consensus consistency target $\hat{x}_0$. One of the two perturbed states is randomly selected for the direct student prediction input $x_t$, and its discrepancy with the consensus target defines the normalized local uncertainty map $\tilde{U}$, which reweights the consistency distillation loss to attenuate supervision at regions with higher uncertainty. A feature-space discriminator provides adversarial and feature-matching objectives, together with a semantic alignment loss.}
 \label{fig:f}
 \vspace{-.2in}
\end{figure*}

\subsection{Preliminary}
\noindent\textbf{Video Diffusion Model.}
Flow matching models learn a vector field $v_\psi(x_t, t; c)$ that transports noise $x_1 \sim \mathcal{N}(0, I)$ to data $x_0$ via the ODE $\text{d}x_t = v_\psi(x_t, t; c)\,\text{d}t$. The model is trained to minimize $\mathbb{E}_{t, x_0, \epsilon}\left[\|v_\psi(x_t, t; c) - (\epsilon - x_0)\|^2\right]$ with $x_t = (1-\sigma_t)x_0 + \sigma_t\epsilon$. High-quality video synthesis requires solving this ODE with tens to hundreds of iterative solver steps, constituting the primary computational bottleneck for deployment.

\noindent\textbf{Consistency Distillation.}
Consistency distillation aims to learn a student model $f_\phi(x_t, t; c)$ that maps any point $x_t$ on the probability flow ODE trajectory directly to the clean endpoint $x_0$ in a single forward pass. The model satisfies the self-consistency property $f_\phi(x_t, t; c) = f_\phi(x_{t'}, t'; c)$ for all $t, t'$ lying on the same trajectory. In practice, this property is enforced along a teacher-guided trajectory: the frozen teacher first performs one Euler step with step size $\Delta t$, moving from $t$ to $t'=t-\Delta t$ to obtain an intermediate state $x_{t'}$. The student then predicts the endpoint from $x_{t'}$ to form the distillation target, while also predicting the endpoint directly from $x_t$. The consistency objective encourages these two student predictions to agree, with the target prediction detached from gradient computation.

\subsection{Uncertainty-Aware Consistency Distillation}
\noindent\textbf{Dual-Path Target Construction.}
To estimate local consistency uncertainty without introducing an additional uncertainty model, we construct two independently perturbed teacher-guided consistency paths. Given a clean latent $x_0$, we randomly sample a timestep $t$ and two independent noise perturbations $\epsilon_1, \epsilon_2 \sim \mathcal{N}(0, I)$ to construct two noisy states:
\begin{equation}
    x_t^{(1)} = \sigma_t \epsilon_1 + (1 - \sigma_t) x_0, \quad
    x_t^{(2)} = \sigma_t \epsilon_2 + (1 - \sigma_t) x_0,
\end{equation}
where $\sigma_t$ denotes the noise level associated with timestep $t$~\citep{lipman2023flow}. For each perturbed input, the teacher performs a one-step Euler update $g_\theta(\cdot)$ from timestep $t$ to $t'=t-\Delta t$ to obtain an intermediate point, which is then fed to the student $f_\phi(\cdot)$ to predict the endpoint. For notational simplicity, we omit the conditioning variable $c$ throughout this section.
Averaging these two predictions yields a consensus consistency target:
\begin{equation}
    \hat{x}_0 = \frac{1}{2}\left(f_\phi(g_\theta(x_t^{(1)}, t),t') +
    f_\phi(g_\theta(x_t^{(2)}, t),t')\right),
\end{equation}
where the two student predictions are computed without gradient tracking.
Independent noise sampling provides two stochastic perturbations of the same underlying latent, allowing us to probe the stability of the resulting teacher-guided consistency predictions. The consensus target aggregates the two independently constructed predictions, reducing its dependence on any individual perturbed path. In parallel, we randomly select one of the two perturbed states, $x_t^{(1)}$ or $x_t^{(2)}$, as the input to the direct student prediction, denoted as $x_t$. The resulting prediction $f_\phi(x_t,t)$ remains fully trainable and is optimized to match the detached consensus target. This shared input ensures that the direct prediction and one of the teacher-guided predictions originate from the same perturbed state, while the other path provides an independent prediction for consensus estimation. Because the consensus target is formed from the teacher-advanced states while the direct prediction starts from the noisy state, the discrepancy also reflects the sensitivity of the student's endpoint to the teacher's single Euler step, in addition to the injected perturbation.

\noindent\textbf{Uncertainty-Aware Loss.}
Following prior discrepancy-based uncertainty estimation methods~\citep{zheng2021rectifying,zhang2025ctrl}, we use the student--consensus discrepancy as a parameter-free proxy for local uncertainty. Specifically, we compute the element-wise squared discrepancy between the student's direct prediction $f_\phi(x_t,t)$ and the consensus target $\hat{x}_0$:
\begin{equation}
U = \mathrm{sg}\left[\left(f_\phi(x_t, t) - \hat{x}_0\right)^2\right],
\end{equation}
where $\mathrm{sg}[\cdot]$ denotes stop-gradient. The resulting uncertainty has the same spatiotemporal latent structure as the student prediction, providing fine-grained uncertainty estimates over spatiotemporal locations. The uncertainty is treated as fixed when optimizing the student within each update. Large values of $U$ indicate regions where the direct student prediction disagrees with the consensus prediction induced by the two teacher-guided paths, suggesting that the corresponding consistency supervision is less reliable. We normalize $U$ by its mean value to obtain $\tilde{U}=U/(\bar{U}+\delta)$ before applying the exponential weight, where $\bar{U}$ is the per-video mean over all spatiotemporal locations, and $\delta$ is a small constant for numerical stability. The uncertainty-aware consistency loss is then defined as:
\begin{equation}
    \mathcal{L}_{\text{ua}} = \mathbb{E}\left[ e^{-\lambda\tilde{U}} \cdot
    \sqrt{(f_\phi(x_t, t) - \hat{x}_0)^2 + \eta} \right],
\end{equation}
where $\lambda$ controls the uncertainty weighting strength, and $\eta$ is a small constant for numerical stability. Since $\tilde{U} \geq 0$ and $\lambda>0$, the exponential weight $e^{-\lambda\tilde{U}} \in (0,1]$ ensures that low-uncertainty spatiotemporal locations retain relatively stronger gradient contributions, while high-uncertainty locations are suppressed.

\noindent\textbf{Why Not a Learnable Uncertainty Model?}
A natural alternative is to introduce a learnable uncertainty head to predict local uncertainty from student features. However, without explicit uncertainty supervision, such a head must learn the notion of uncertainty indirectly through the distillation objective, making its estimates sensitive to optimization dynamics and potentially poorly aligned with actual consistency errors. We instead derive uncertainty from the consistency discrepancy, directly linking uncertainty to the reliability of the distillation signal. Our experiments in Sec.~\ref{sec:ablation} and App.~\ref{sec:a} show that two perturbed paths are sufficient for uncertainty estimation.

\subsection{Adversarial Refinement with Semantic Alignment}
Training with the consistency loss alone tends to produce blurry outputs due to regression toward the mean. We complement it with two auxiliary objectives to preserve perceptual quality during aggressive step reduction.

\noindent\textbf{Feature-Space Adversarial Loss.}
Following prior distillation works~\citep{dcm,lu2025adversarial}, we employ a discriminator $\mathcal{D}$ on intermediate transformer features, which provide richer structural and semantic information than pixel space. Specifically, $h_{\mathrm{fake}}$ and $h_{\mathrm{real}}$ are extracted by the teacher transformer from student-generated samples and corresponding real samples, respectively, and then fed to $\mathcal{D}$. The generator and discriminator losses are:
\begin{align}
\mathcal{L} _{G}^{\text{adv}}=\mathbb{E} \left[ \max \left( 0,1-\mathcal{D} \left( h_{fake} \right) \right) +\lambda _{\text{feat}}\parallel h_{real}-h_{fake}
\parallel _{2}^{2} \right] ,
\\
\mathcal{L} _{D}^{\text{adv}}=\mathbb{E} \left[ \max \left( 0,1-\mathcal{D} \left( h_{real} \right)
\right) +\max \left( 0,1+\mathcal{D} \left( h_{fake} \right) \right) \right] ,
\end{align}
where $\lambda_{\text{feat}}$ balances the feature matching term.

\noindent\textbf{Semantic Alignment with Frozen Vision Embeddings.}
To provide semantic guidance for adversarial training, we align intermediate discriminator representations with frozen DINOv2~\citep{dinov2} representations extracted from real video frames. Specifically, the selected teacher features are processed by the corresponding discriminator heads and projected by a learnable head $\mathrm{Proj}(\cdot)$ to match the dimension of $f_{\text{vis}}(\cdot)$. The alignment loss is:
\begin{equation}
    \mathcal{L}_{\text{align}} =
    -\mathbb{E}\left[\cos(\mathrm{Proj}(\mathcal{D}(h_{real})),f_{\text{vis}}(y))\right],
\end{equation}
where $y$ denotes the corresponding ground-truth video frames temporally subsampled to match the latent temporal resolution. The alignment provides semantic guidance to the adversarial signal and helps reduce content drift during few-step generation.

\subsection{Training Procedure}
As shown in Figure~\ref{fig:f},  we train $f_\phi$ with LoRA adapters~\citep{lora} on frozen base weights, while jointly optimizing $\mathcal{D}$ and $\mathrm{Proj}$.  %The student model and the discriminator are optimized with separate objectives. 
The total loss for the student model $f_\phi$ combines the uncertainty-aware distillation loss and the adversarial generator loss, while the discriminator $\mathcal{D}$ is trained with the adversarial loss augmented by the semantic alignment loss:
\begin{equation}
    \mathcal{L}_{S} = \mathcal{L}_{\text{ua}} + \lambda_{\text{adv}} \mathcal{L}_{G}^{\text{adv}},\quad    \mathcal{L}_{D} = \mathcal{L}_{D}^{\text{adv}} + \lambda_{\text{align}} \, \mathcal{L}_{\text{align}},
\end{equation}
where $\lambda_{\text{adv}}$ is the adversarial loss weight, and $\lambda_{\text{align}}$ controls the strength of semantic regularization. 

\section{Experiment}

\noindent\textbf{Setting.} We use Wan2.1-T2V-1.3B~\citep{wan} as the teacher model for distillation. The student reuses the teacher weights and is equipped with LoRA adapters (rank 128). We perform distillation on 81-frame video sequences at a resolution of $832 \times 480$, using a batch size of 4. The student and discriminator are optimized via AdamW~\citep{adamw} with learning rates of $4\!\times\!10^{-5}$ and $1\!\times\!10^{-5}$, respectively. Semantic alignment uses a frozen DINOv2 ViT-B/14 encoder. The uncertainty weighting strength is set to $\lambda=1$, and the semantic alignment loss weight is set to $\lambda_{\text{align}}=1$. We provide ablation studies on these two hyperparameters in App.~\ref{sec:b} and App.~\ref{sec:c}, respectively. We follow the experimental protocol of DCM~\citep{dcm} for our distillation framework, setting the adversarial loss weight $\lambda_{\text{adv}}=0.5$ and the feature-matching loss weight $\lambda_{\text{feat}}=1$. All experiments are conducted on a single NVIDIA H800 80GB GPU. We distill for approximately 2,000 steps, which takes about 20 hours to complete.

\noindent\textbf{Evaluation Metrics.}
For video quality evaluation, we adopt VBench 2.0~\citep{vb2} as our primary metric, which provides a fixed set of prompts and standardized evaluation protocols for assessing video generation quality. It systematically evaluates adherence to real-world physical laws and semantics across 18 fine-grained sub-abilities, aggregated into five key dimensions: Human Fidelity, Controllability, Creativity, Physics, and Commonsense. We follow the benchmark's fixed prompt set and evaluate generated videos under multiple random seeds to ensure robust and reproducible comparisons. To complement automatic metrics, we further conduct a user study on subjective visual quality and semantic coherence.

\subsection{Comparison with Competitive Methods}
\noindent\textbf{Competitive Methods.} We compare against five representative methods spanning both multi-step video generation models and recent distillation approaches. \textit{Wan2.1-1.3B}~\citep{wan} and \textit{CogVideoX-5B}~\citep{cogvideox} are included as multi-step (50-step) baselines that provide reference points for generation quality under substantially more inference steps. For distillation-based methods, we include three state-of-the-art approaches: \textit{DCM}~\citep{dcm} employs a staged distillation strategy based on the \textit{Wan2.1-1.3B} teacher and enables complete video generation within 4 inference steps; \textit{CausalForcing}~\citep{causal} and \textit{OneForcing}~\citep{oneforcing} use \textit{Wan2.1-14B} as the teacher and adopt autoregressive frame-wise distillation. CausalForcing achieves 4-step generation for all frames, while OneForcing applies 4 steps only to the first frame and reduces subsequent frames to single-step inference, trading off initial fidelity for faster sequential generation.

\noindent\textbf{Quantitative Comparison.}
As shown in Table~\ref{tab:t1}, our method achieves the highest VBench 2.0 mean score of \textbf{0.556} with only 4 inference steps using a lightweight 1.3B teacher, surpassing both 50-step baselines and all distillation baselines. Compared to Wan2.1-1.3B and CogVideoX-5B, which require 50 steps, our approach delivers substantial improvements in all metrics while reducing computational cost by over an order of magnitude. Among distillation methods, our method leads in Human Fidelity (0.861), Creativity (0.642), and Physics (0.516). The high Creativity score suggests that the proposed distillation framework can preserve diverse generation under aggressive step reduction. OneForcing and DCM are marginally better in Controllability and Commonsense but worse elsewhere. These results indicate that our consistency distillation framework with dual-path target prediction and discrepancy-based uncertainty estimation enables a 1.3B student to match or exceed the quality of larger models and competitive distillation baselines under highly efficient inference conditions.

\begin{figure*}[t]
 \centering
 %\vspace{-.25in}
 \includegraphics[width=\linewidth]{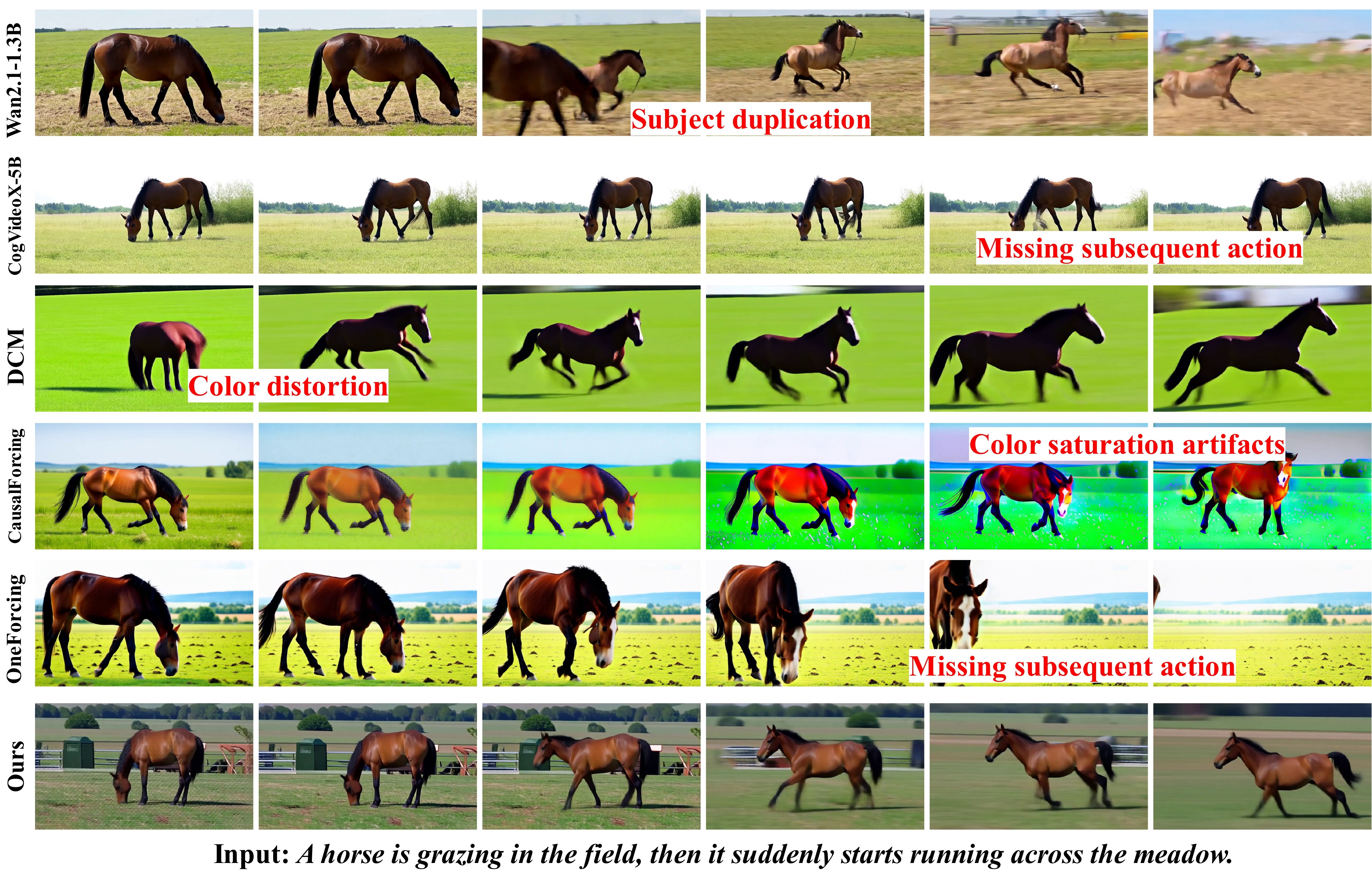}
 \vspace{-.35in}
 \caption{Qualitative comparison with competitive methods. Our method generates videos with superior visual quality, accurate temporal transitions, and full prompt adherence compared to all baselines, highlighting that our uncertainty modeling enables the student to surpass its own teacher. More sample videos are available on our anonymous project website.}
 \vspace{-.2in}
 \label{fig:base}
\end{figure*}

\noindent\textbf{Qualitative Comparison.} As shown in Figure~\ref{fig:base}, we compare representative outputs for the prompt ``\textit{A horse is grazing in the field, then it suddenly starts running across the meadow}''. Wan2.1-1.3B generates two distinct horses for the grazing and running actions, breaking entity continuity even though both motions are present. CogVideoX-5B, CausalForcing, and OneForcing capture the initial grazing scene but omit the subsequent running motion, neglecting the specified temporal transition. CausalForcing further exhibits progressive color drift, which may be associated with error accumulation in autoregressive generation. DCM exhibits noticeable color distortion and less natural textures in our comparison. Notably, our method adopts Wan2.1-1.3B itself as the teacher model, yet generates videos of higher visual quality and stronger prompt alignment than the teacher. Our uncertainty-aware distillation relaxes the supervision on unreliable consistency constraints relative to reliable ones, enabling the student to further improve visual fidelity and temporal-semantic coherence beyond the teacher's output.

\begin{table}[t]
  \centering
  \vspace{-.1in}  
  \caption{Quantitative comparison on VBench 2.0 across 5 video quality dimensions. All videos are generated at $832\times480$ resolution with $\ell=81$ frames. NFE denotes the number of function evaluations during inference. Our method achieves the highest mean score with 4 NFEs using a 1.3B teacher. DCM also uses a 1.3B teacher, while CausalForcing and OneForcing use a 14B teacher.}
  \resizebox{0.97\linewidth}{!}{
    \begin{tabular}{c|c|c|ccccc|c}
    \toprule
    Method & NFE  & Teacher & Human Fidelity & Creativity & Controllability & Commonsense & Physics & Mean \\
    \midrule
    Wan2.1-1.3B & 50    &  -    & 0.711  & 0.551  & 0.099  & 0.519  & 0.382  & 0.452  \\
    CogVideoX-5B & 50    & -     & 0.797  & 0.344  & 0.199  & 0.539  & 0.405  & 0.457  \\
    \midrule
    \rowcolor{gray!15}
    \multicolumn{9}{c}{\textbf{Distillation Methods}} \\
    CausalForcing & $4+4\times(\ell//4)$     & Wan2.1-14B & 0.855  & 0.462  & 0.184  & 0.565  & 0.492  & 0.512  \\
    OneForcing & $4+1\times(\ell//4)$     & Wan2.1-14B & 0.754  & 0.631  & \textbf{0.213} & 0.571  & 0.460  & 0.526  \\
    DCM   & 4     & Wan2.1-1.3B & 0.828  & 0.599  & 0.209  & \textbf{0.603} & 0.394  & 0.527  \\
    UACD(Ours)  & 4     & Wan2.1-1.3B & \textbf{0.861} & \textbf{0.642} & 0.203  & 0.556  & \textbf{0.516} & \textbf{0.556} \\
    \bottomrule
    \end{tabular}%
    }
  \vspace{-.15in}
  \label{tab:t1}%
\end{table}%

\begin{figure*}[t]
%\vspace{-.1in}
  \centering
  \begin{minipage}[b]{0.48\linewidth}
    \centering
    \includegraphics[width=\linewidth]{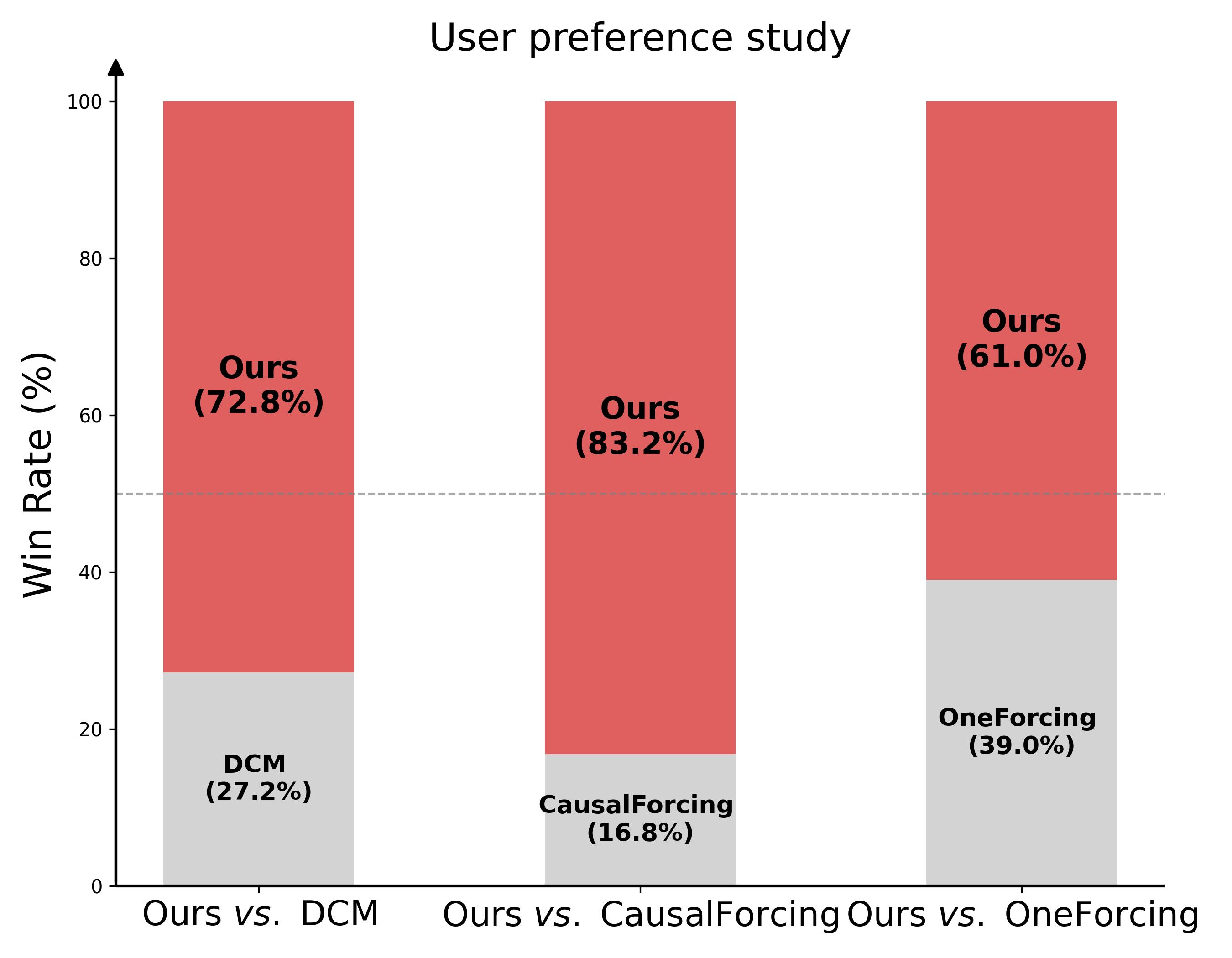}
    \vspace{-.35in}
    \caption{User preference study results. Our approach is favored by a clear majority of users across all pairwise comparisons, reflecting its superior perceptual quality.}
    \vspace{-.2in}
    \label{fig:us}
  \end{minipage}\hfill
  \begin{minipage}[b]{0.48\linewidth}
    \centering
    \includegraphics[width=\linewidth]{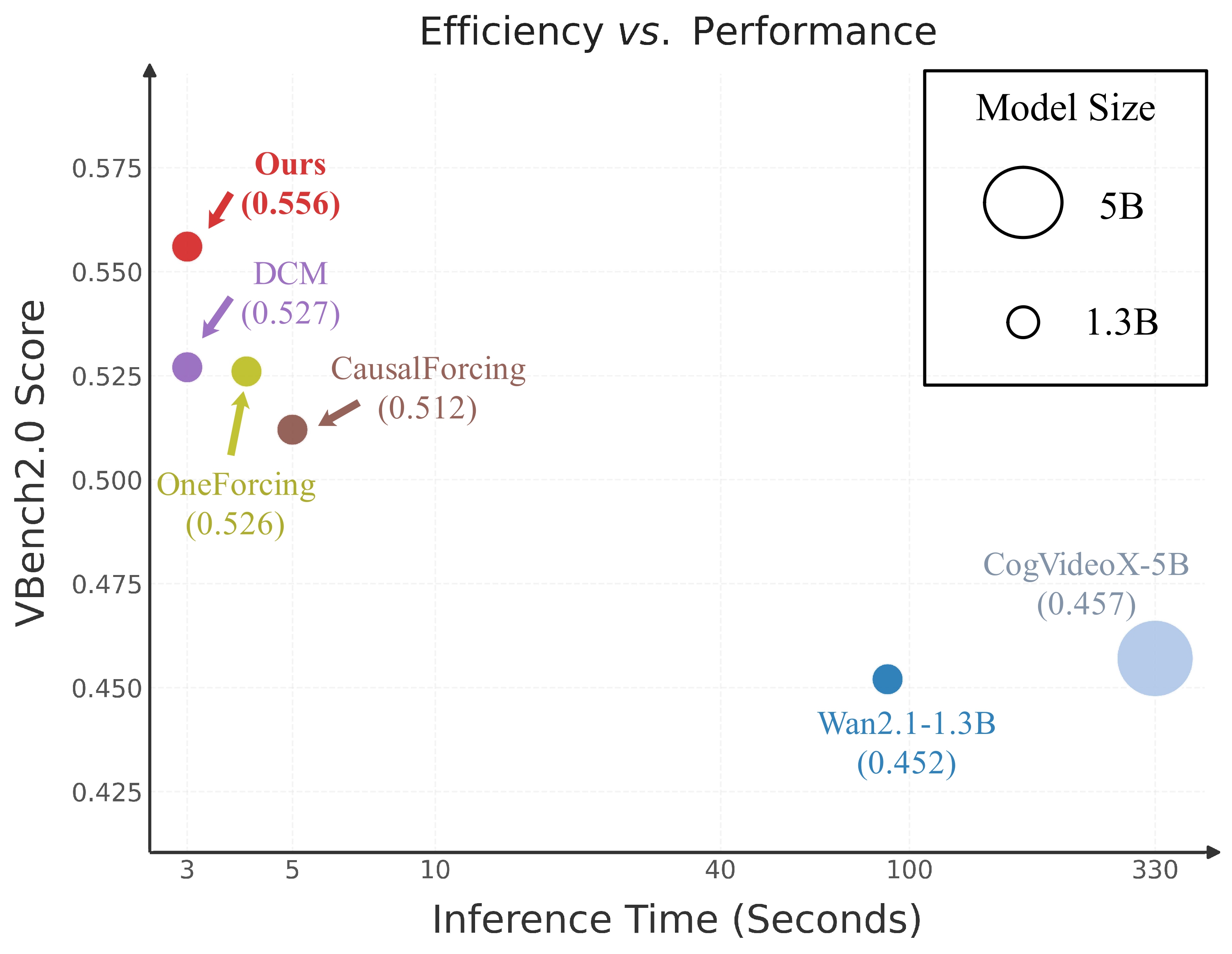}
    \vspace{-.35in}
    \caption{Efficiency comparison across methods. Methods closer to the top-left achieve superior performance, indicating higher generation quality with reduced inference time.}
    \vspace{-.2in}
    \label{fig:eff}
  \end{minipage}
\end{figure*}

\noindent\textbf{Efficiency.}
As shown in Figure~\ref{fig:eff}, our method achieves a favorable balance between generation quality and inference speed among all evaluated approaches. Notably, with only four inference steps, our student model surpasses the generation quality of its teacher (Wan2.1-1.3B) while remaining highly efficient for long video synthesis. Although OneForcing enables one-step generation at the frame level, its autoregressive design still requires sequential generation across 81 frames, whereas our video-level approach requires only four sequential steps for the entire video, avoiding the error accumulation of frame-wise autoregression.

\noindent\textbf{User Study.} Following the human evaluation protocol of prior works~\citep{dcm,hunyuanvideo,vb2}, we randomly sample 30 videos per model. Each trial presents a text prompt and two videos, one from our method and one from a competing distillation approach, with randomized order. Raters select the preferred video based on text alignment, motion quality, and visual quality. Each sample is evaluated by 30 independent raters, with aggregated results shown in Figure~\ref{fig:us}. The voting results indicate that our method is more preferred.

\subsection{Ablation Studies and Further Discussion}
\label{sec:ablation}
To isolate the contribution of each component, we conduct controlled experiments with six configurations: \textit{Variant 1} retains only plain consistency distillation together with basic adversarial training, without semantic alignment or uncertainty modeling; \textit{Variant 2} builds upon Variant 1 by incorporating semantic alignment into the adversarial objective; \textit{Variant 3} uses a single teacher-guided path to construct one teacher-guided supervision target, and estimates uncertainty from the discrepancy between this target and the student's direct prediction; \textit{Variant 4} keeps the dual-path construction but replaces the discrepancy-based uncertainty proxy with a freely learnable uncertainty token optimized end-to-end; \textit{Variant 5} keeps the complete pipeline but removes the semantic alignment from the adversarial objective; and \textit{Variant 6} is our full model.

\noindent\textbf{Quantitative Effect of Primary Components.}
As shown in Table~\ref{tab:ab}, Variant 1 achieves near-perfect Human Fidelity (0.998) but lower scores on Physics (0.169) and Controllability (0.146). Without semantic alignment or uncertainty reweighting, the student produces clean but semantically and physically wrong frames, which Human Fidelity does not penalize since it rewards frame-level realism rather than semantic or physical correctness. The configuration without uncertainty modeling (Variant 2) achieves a mean VBench 2.0 score of only 0.462, whereas introducing the single-path uncertainty proxy (Variant 3) raises it to 0.513, indicating that uncertainty-aware supervision helps to stabilize consistency distillation. Replacing the discrepancy-based proxy with a freely learnable uncertainty token (Variant 4) results in a lower mean score of 0.509, suggesting that directly grounding uncertainty in prediction discrepancies provides a more effective signal than learning uncertainty weights solely from the distillation objective. These results suggest that explicit uncertainty estimation is beneficial for consistency distillation. Moreover, combining the proposed dual-path construction with discrepancy-based uncertainty estimation yields a more effective uncertainty-aware supervision signal than either the single-path proxy or the learnable uncertainty token. We further investigate the number of target paths in the supplementary material. Extending the dual-path construction to three and four paths yields comparable performance at higher computational cost, validating our dual-path design as an efficient choice.

\begin{table}[t]
  \centering
  \caption{Ablation study on primary components. \textbf{CD}: Consistency Distillation baseline with adversarial training. \textbf{DP}: Dual-path Prediction, where two independently perturbed teacher-guided paths are constructed and the student predicts the endpoint from each path. \textbf{UT}: Uncertainty Type. \textbf{ED}: Uncertainty Estimation via Discrepancy, where the discrepancy between the student's direct endpoint prediction and the consensus consistency target is used as a parameter-free uncertainty estimate. \textbf{LT}: Uncertainty Estimation via a Learnable Token. \textbf{DSA}: Discriminator Semantic Alignment loss applied during adversarial training. Variant 1 achieves near-perfect Human Fidelity but a substantially lower mean, since Human Fidelity rewards frame-level realism but not semantic or physical correctness. Our full model (Variant 6) achieves the highest mean with balanced performance, confirming that dual-path uncertainty and semantic alignment are complementary. }
  \resizebox{0.97\linewidth}{!}{
     \begin{tabular}{c|cccc|ccccc|c}
    \toprule
          & \multicolumn{4}{c|}{Variants}         & \multicolumn{5}{c|}{VBench 2.0} & \\
          & CD    & DP    & UT   & DSA   & Human Fidelity & Creativity & Controllability & Commonsense & Physics & Mean \\
    \midrule
    (1)   & \checkmark &       &    &       & \textbf{0.998} & 0.396  & 0.146  & 0.509  & 0.169  & 0.444  \\
    (2)   & \checkmark &       &            & \checkmark & 0.597  & 0.632  & 0.108  & 0.551  & 0.422  & 0.462  \\
    (3)   & \checkmark &       & ED      & \checkmark & 0.840  & 0.584  & 0.197  & 0.487  & 0.456  & 0.513  \\
    (4)   & \checkmark & \checkmark &  LT & \checkmark & 0.812  & 0.605  & 0.153  & 0.516  & 0.458  & 0.509  \\
    (5)   & \checkmark & \checkmark & ED      &       & 0.854  & \textbf{0.682} & 0.142  & 0.551  & 0.422  & 0.530  \\
    (6)   & \checkmark & \checkmark & ED      & \checkmark & 0.861  & 0.642  & \textbf{0.203} & \textbf{0.556} & \textbf{0.516} & \textbf{0.556} \\
    \bottomrule
    \end{tabular}%
    }
  \label{tab:ab}%
  %\vspace{-.2in}
\end{table}%

\begin{figure*}[t]
%\vspace{-.2in}
 \centering
 \includegraphics[width=\linewidth]{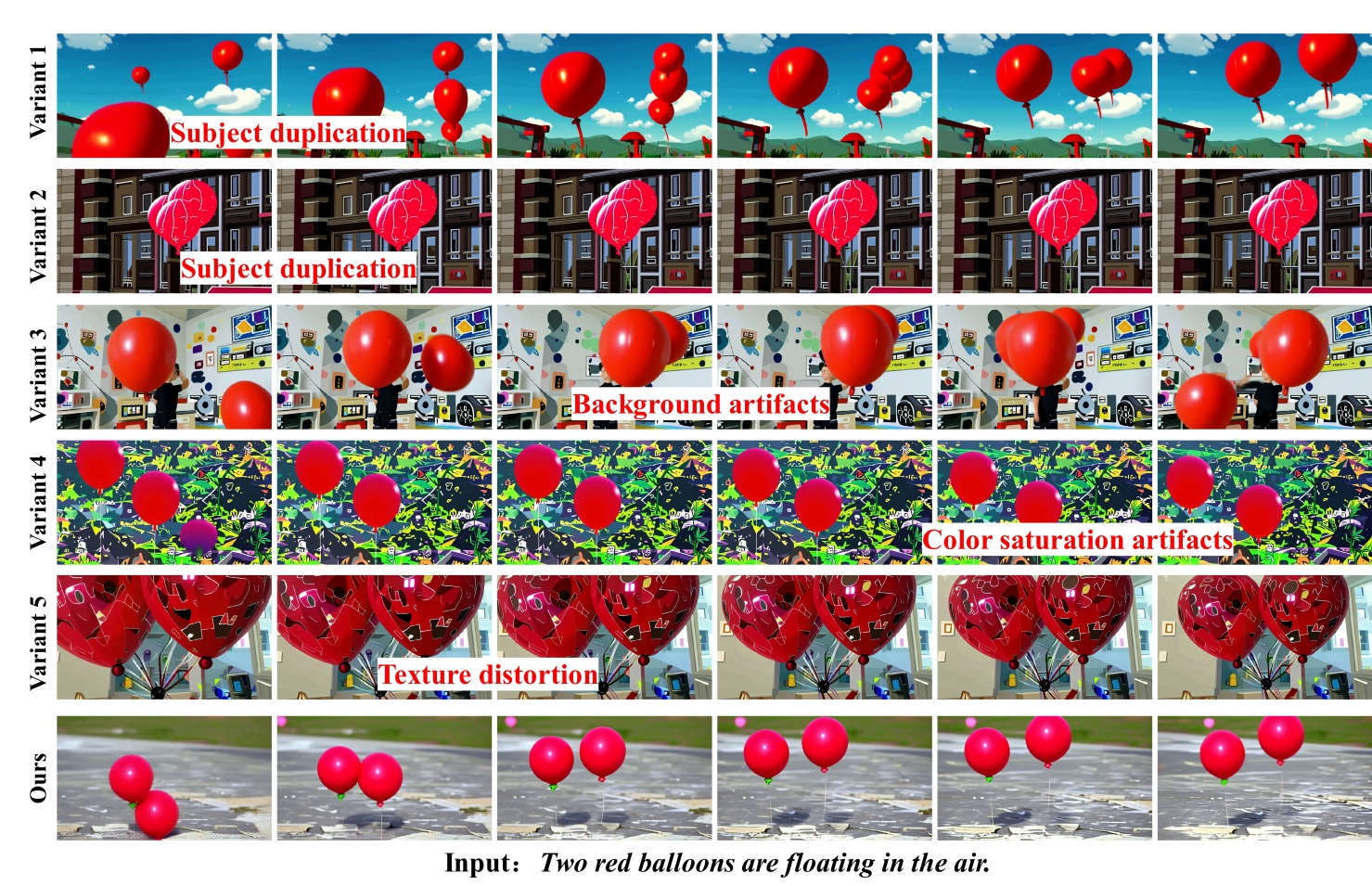}
 \vspace{-.25in}
 \caption{Qualitative comparison of ablation variants. Our full model generates videos that faithfully align with the text prompt and closely resemble real-world video content in motion dynamics and visual fidelity.}
 \label{fig:ab}
 \vspace{-.2in}
\end{figure*}

\noindent\textbf{Qualitative Effect of Primary Components.} As shown in Figure~\ref{fig:ab}, we visualize ablation results under the prompt ``\textit{Two red balloons are floating in the air.}'' to examine how each component affects generation quality. Variants~1 and~2 produce an incorrect number of balloons, and Variant~1 further yields color tones that deviate markedly from real-world videos, indicating that plain consistency distillation without semantic alignment struggles with both object counting and photorealism. Introducing uncertainty estimation via single-path target (Variant~3) or a learnable token (Variant~4) leads to cluttered background scenes, with Variant~4 additionally exhibiting excessively saturated background colors, suggesting that these alternative uncertainty estimation strategies are less effective at preserving clean spatial structure. Removing semantic alignment from the adversarial objective (Variant~5) yields distorted balloon textures, confirming that semantic-aligned adversarial training is essential for preserving fine-grained structural fidelity. Only our full model simultaneously achieves accurate prompt adherence, natural color reproduction, clean background composition, and realistic motion, suggesting that all proposed components are complementary and jointly beneficial for high-quality few-step video generation.

\noindent\textbf{Limitations.} Our method achieves high-quality video generation in 4 inference steps, but extending consistency distillation to 1-step remains difficult. As discussed in OneForcing~\citep{oneforcing}, the sharply concentrated curvature of video teacher trajectories near the high-noise endpoint degrades trajectory-based consistency objectives when the trajectory is compressed into a single step. 

\section{Conclusion}
We introduce Uncertainty-Aware Consistency Distillation for high-quality few-step video generation.
Our approach recognizes that consistency supervision is not equally reliable across spatiotemporal locations, with its reliability varying according to how much the content changes over time. We derive a parameter-free local uncertainty signal from independently perturbed teacher-guided paths to adapt the strength of distillation supervision.
Combined with semantic-aligned adversarial training, our method enables effective distillation into four inference steps, surpassing both the teacher and existing distillation methods on VBench 2.0 and human evaluations. These results show the effectiveness of uncertainty-aware supervision for improving the fidelity and semantic coherence of few-step video generation.

\section{AI use statement}
In this work, we used generative AI tools (\eg, ChatGPT) to edit this paper to improve readability, specifically for grammar and spelling checks. We have not used generative AI tools for creating scientific figures or for any content that would constitute plagiarism. We have carefully reviewed all AI-assisted text revisions to ensure their accuracy. We take full responsibility for the final content of this work.

\bibliography{conference}

@article{wan,
  title={Wan: Open and advanced large-scale video generative models},
  author={Wan, Team and Wang, Ang and Ai, Baole and Wen, Bin and Mao, Chaojie and Xie, Chen-Wei and Chen, Di and Yu, Feiwu and Zhao, Haiming and Yang, Jianxiao and others},
  journal={arXiv},
  year={2025}
}

@article{hunyuanvideo,
  title={Hunyuanvideo: A systematic framework for large video generative models},
  author={Kong, Weijie and Tian, Qi and Zhang, Zijian and Min, Rox and Dai, Zuozhuo and Zhou, Jin and Xiong, Jiangfeng and Li, Xin and Wu, Bo and Zhang, Jianwei and others},
  journal={arXiv},
  year={2024}
}

@inproceedings{cogvideox,
  title={Cogvideox: Text-to-video diffusion models with an expert transformer},
  author={Yang, Zhuoyi and Teng, Jiayan and Zheng, Wendi and Ding, Ming and Huang, Shiyu and Xu, Jiazheng and Yang, Yuanming and Hong, Wenyi and Zhang, Xiaohan and Feng, Guanyu and others},
  booktitle={International Conference on Learning Representations},
  volume={2025},
  pages={83048--83077},
  year={2025}
}

@article{sora,
  title={Sora: A review on background, technology, limitations, and opportunities of large vision models},
  author={Liu, Yixin and Zhang, Kai and Li, Yuan and Yan, Zhiling and Gao, Chujie and Chen, Ruoxi and Yuan, Zhengqing and Huang, Yue and Sun, Hanchi and Gao, Jianfeng and others},
  journal={arXiv preprint arXiv:2402.17177},
  year={2024}
}

@article{song2020denoising,
  title={Denoising diffusion implicit models},
  author={Song, Jiaming and Meng, Chenlin and Ermon, Stefano},
  journal={arXiv preprint arXiv:2010.02502},
  year={2020}
}

@inproceedings{meng2023distillation,
  title={On distillation of guided diffusion models},
  author={Meng, Chenlin and Rombach, Robin and Gao, Ruiqi and Kingma, Diederik and Ermon, Stefano and Ho, Jonathan and Salimans, Tim},
  booktitle={2023 IEEE/CVF Conference on Computer Vision and Pattern Recognition (CVPR)},
  pages={14297--14306},
  year={2023},
  organization={IEEE}
}

@inproceedings{gu2023boot,
  title={Boot: Data-free distillation of denoising diffusion models with bootstrapping},
  author={Gu, Jiatao and Zhai, Shuangfei and Zhang, Yizhe and Liu, Lingjie and Susskind, Joshua M},
  booktitle={ICML 2023 Workshop on Structured Probabilistic Inference \& Generative Modeling},
  year={2023}
}

@inproceedings{dcm,
  title={Dual-expert consistency model for efficient and high-quality video generation},
  author={Lv, Zhengyao and Si, Chenyang and Pan, Tianlin and Chen, Zhaoxi and Wong, Kwan-Yee K and Qiao, Yu and Liu, Ziwei},
  booktitle={2025 IEEE/CVF International Conference on Computer Vision (ICCV)},
  pages={14983--14993},
  year={2025},
  organization={IEEE}
}

@article{oneforcing,
  title={One-Forcing: Towards Stable One-Step Autoregressive Video Generation},
  author={Feng, Jiaqi and Cui, Justin and Ban, Yuanhao and Hsieh, Cho-Jui},
  journal={arXiv preprint arXiv:2605.23458},
  year={2026}
}

@inproceedings{causal,
title={Causal Forcing: Autoregressive Diffusion Distillation Done Right for High-Quality Real-Time Interactive Video Generation},
author={Hongzhou Zhu and Min Zhao and Guande He and Hang Su and Chongxuan Li and Jun Zhu},
booktitle={Forty-third International Conference on Machine Learning},
year={2026},
url={https://openreview.net/forum?id=BYInOck3gr}
}

@article{dinov2,
  title={Dinov2: Learning robust visual features without supervision},
  author={Oquab, Maxime and Darcet, Timoth{\'e}e and Moutakanni, Th{\'e}o and Vo, Huy and Szafraniec, Marc and Khalidov, Vasil and Fernandez, Pierre and Haziza, Daniel and Massa, Francisco and El-Nouby, Alaaeldin and others},
  journal={arXiv preprint arXiv:2304.07193},
  year={2023}
}

@article{lora,
  title={Lora: Low-rank adaptation of large language models},
  author={Hu, Edward J and Shen, Yelong and Wallis, Phillip and Allen-Zhu, Zeyuan and Li, Yuanzhi and Wang, Shean and Wang, Lu and Chen, Weizhu},
  journal={arXiv preprint arXiv:2106.09685},
  year={2021}
}

@article{vb2,
  title={Vbench-2.0: Advancing video generation benchmark suite for intrinsic faithfulness},
  author={Zheng, Dian and Huang, Ziqi and Liu, Hongbo and Zou, Kai and He, Yinan and Zhang, Fan and Gu, Lulu and Zhang, Yuanhan and He, Jingwen and Zheng, Wei-Shi and others},
  journal={arXiv preprint arXiv:2503.21755},
  year={2025}
}

@inproceedings{adamw,
title={Decoupled Weight Decay Regularization},
author={Ilya Loshchilov and Frank Hutter},
booktitle={International Conference on Learning Representations},
year={2019},
url={https://openreview.net/forum?id=Bkg6RiCqY7},
}

@inproceedings{lu2025adversarial,
  title={Adversarial distribution matching for diffusion distillation towards efficient image and video synthesis},
  author={Lu, Yanzuo and Ren, Yuxi and Xia, Xin and Lin, Shanchuan and Wang, Xing and Xiao, Xuefeng and Ma, Andy J and Xie, Xiaohua and Lai, Jian-Huang},
  booktitle={2025 IEEE/CVF International Conference on Computer Vision (ICCV)},
  pages={16818--16829},
  year={2025},
  organization={IEEE}
}

@inproceedings{Song2023Consistency,
author = {Song, Yang and Dhariwal, Prafulla and Chen, Mark and Sutskever, Ilya},
title = {Consistency models},
year = {2023},
publisher = {JMLR.org},
booktitle = {Proceedings of the 40th International Conference on Machine Learning},
articleno = {1335},
numpages = {42},
location = {Honolulu, Hawaii, USA},
series = {ICML'23}
}

@article{luo2024latent,
  title={Latent consistency models: Synthesizing high-resolution images with few-step inference},
  author={Luo, Simian and Tan, Yiqin and Huang, Longbo and Li, Jian and Zhao, Hang},
  journal={arXiv preprint arXiv:2310.04378},
  year={2023}
}

@article{wang2024phased,
  title={Phased consistency models},
  author={Wang, Fu-Yun and Huang, Zhaoyang and Bergman, Alexander W and Shen, Dazhong and Gao, Peng and Lingelbach, Michael and Sun, Keqiang and Bian, Weikang and Song, Guanglu and Liu, Yu and others},
  journal={Advances in neural information processing systems},
  volume={37},
  pages={83951--84009},
  year={2024}
}

@inproceedings{salimans2022progressive,
title={Progressive Distillation for Fast Sampling of Diffusion Models},
author={Tim Salimans and Jonathan Ho},
booktitle={International Conference on Learning Representations},
year={2022},
url={https://openreview.net/forum?id=TIdIXIpzhoI}
}

@inproceedings{sauer2024adversarial,
  title={Adversarial diffusion distillation},
  author={Sauer, Axel and Lorenz, Dominik and Blattmann, Andreas and Rombach, Robin},
  booktitle={European Conference on Computer Vision},
  pages={87--103},
  year={2024},
  organization={Springer}
}

@inproceedings{
wang2023diffusiongan,
title={Diffusion-{GAN}: Training {GAN}s with Diffusion},
author={Zhendong Wang and Huangjie Zheng and Pengcheng He and Weizhu Chen and Mingyuan Zhou},
booktitle={The Eleventh International Conference on Learning Representations },
year={2023},
url={https://openreview.net/forum?id=HZf7UbpWHuA}
}

@inproceedings{yin2024one,
  title={One-step diffusion with distribution matching distillation},
  author={Yin, Tianwei and Gharbi, Micha{\"e}l and Zhang, Richard and Shechtman, Eli and Durand, Fredo and Freeman, William T and Park, Taesung},
  booktitle={2024 IEEE/CVF Conference on Computer Vision and Pattern Recognition (CVPR)},
  pages={6613--6623},
  year={2024},
  organization={IEEE}
}

@InProceedings{Chen_2026_CVPR,
    author    = {Chen, Guanjie and Huang, Shirui and Sun, Yifu and Liu, Kai and Zhu, Jianchen and Qu, Xiaoye and Cheng, Yu and Chen, Peng},
    title     = {Flash-DMD: Towards High-Fidelity Few-Step Image Generation with Efficient Distillation and Joint Reinforcement Learning},
    booktitle = {Proceedings of the IEEE/CVF Conference on Computer Vision and Pattern Recognition (CVPR)},
    month     = {June},
    year      = {2026},
    pages     = {6010-6020}
}

@article{kendall2017uncertainties,
  title={What uncertainties do we need in bayesian deep learning for computer vision?},
  author={Kendall, Alex and Gal, Yarin},
  journal={Advances in neural information processing systems},
  volume={30},
  year={2017}
}

@article{wang2026rigi,
  title={Rigi: Rectifying image-to-3d generation inconsistency via uncertainty-aware learning},
  author={Wang, Jiacheng and Zheng, Zhedong and Xu, Wei and Liu, Ping},
  journal={IEEE Transactions on Image Processing},
  year={2026},
  publisher={IEEE}
}

@inproceedings{chen2024composed,
  title={Composed image retrieval with text feedback via multi-grained uncertainty regularization},
  author={Chen, Yiyang and Zheng, Zhedong and Ji, Wei and Qu, Leigang and Chua, Tat-Seng},
  booktitle={International Conference on Learning Representations},
  volume={2024},
  pages={54915--54927},
  year={2024}
}

@inproceedings{zhang2025ctrl,
  title={Ctrl-u: Robust conditional image generation via uncertainty-aware reward modeling},
  author={Zhang, Guiyu and Gao, Huan-ang and Jiang, Zijian and Zhao, Hao and Zheng, Zhedong},
  booktitle={International Conference on Learning Representations},
  volume={2025},
  pages={86480--86499},
  year={2025}
}

@inproceedings{jiang2024fisherrf,
  title={Fisherrf: Active view selection and mapping with radiance fields using fisher information},
  author={Jiang, Wen and Lei, Boshu and Daniilidis, Kostas},
  booktitle={European conference on computer vision},
  pages={422--440},
  year={2024},
  organization={Springer}
}

@inproceedings{blattmann2023align,
  title={Align your latents: High-resolution video synthesis with latent diffusion models},
  author={Blattmann, Andreas and Rombach, Robin and Ling, Huan and Dockhorn, Tim and Kim, Seung Wook and Fidler, Sanja and Kreis, Karsten},
  booktitle={2023 IEEE/CVF Conference on Computer Vision and Pattern Recognition (CVPR)},
  pages={22563--22575},
  year={2023},
  organization={IEEE}
}

@article{ho2020denoising,
  title={Denoising diffusion probabilistic models},
  author={Ho, Jonathan and Jain, Ajay and Abbeel, Pieter},
  journal={Advances in neural information processing systems},
  volume={33},
  pages={6840--6851},
  year={2020}
}

@article{ho2022video,
  title={Video diffusion models},
  author={Ho, Jonathan and Salimans, Tim and Gritsenko, Alexey and Chan, William and Norouzi, Mohammad and Fleet, David J},
  journal={Advances in neural information processing systems},
  volume={35},
  pages={8633--8646},
  year={2022}
}

@article{mansourian2025comprehensive,
  title={A comprehensive survey on knowledge distillation},
  author={Mansourian, Amir M and Ahmadi, Rozhan and Ghafouri, Masoud and Babaei, Amir Mohammad and Golezani, Elaheh Badali and Ghamchi, Zeynab Yasamani and Ramezanian, Vida and Taherian, Alireza and Dinashi, Kimia and Miri, Amirali and others},
  journal={arXiv preprint arXiv:2503.12067},
  year={2025}
}

@article{luhman2021knowledge,
  title={Knowledge distillation in iterative generative models for improved sampling speed},
  author={Luhman, Eric and Luhman, Troy},
  journal={arXiv preprint arXiv:2101.02388},
  year={2021}
}

@inproceedings{zheng2023fast,
  title={Fast sampling of diffusion models via operator learning},
  author={Zheng, Hongkai and Nie, Weili and Vahdat, Arash and Azizzadenesheli, Kamyar and Anandkumar, Anima},
  booktitle={International conference on machine learning},
  pages={42390--42402},
  year={2023},
  organization={PMLR}
}

@inproceedings{frans2025one,
  title={One step diffusion via shortcut models},
  author={Frans, Kevin and Hafner, Danijar and Levine, Sergey and Abbeel, Pieter},
  booktitle={International Conference on Learning Representations},
  volume={2025},
  pages={34668--34684},
  year={2025}
}

@inproceedings{ding2025dollar,
  title={Dollar: Few-step video generation via distillation and latent reward optimization},
  author={Ding, Zihan and Jin, Chi and Liu, Difan and Zheng, Haitian and Singh, Krishna Kumar and Zhang, Qiang and Kang, Yan and Lin, Zhe and Liu, Yuchen},
  booktitle={2025 IEEE/CVF International Conference on Computer Vision (ICCV)},
  pages={17961--17971},
  year={2025},
  organization={IEEE}
}

@inproceedings{yang2026towards,
  title={Towards one-step causal video generation via adversarial self-distillation},
  author={Yang, Yongqi and Huang, Huayang and Peng, Xu and Hu, Xiaobin and Luo, Donghao and Zhang, Jiangning and Wang, Chengjie and Wu, Yu},
  booktitle={International Conference on Learning Representations},
  volume={2026},
  pages={34858--34876},
  year={2026}
}

@inproceedings{raghu2019direct,
  title={Direct uncertainty prediction for medical second opinions},
  author={Raghu, Maithra and Blumer, Katy and Sayres, Rory and Obermeyer, Ziad and Kleinberg, Bobby and Mullainathan, Sendhil and Kleinberg, Jon},
  booktitle={International conference on machine learning},
  pages={5281--5290},
  year={2019},
  organization={PMLR}
}

@article{nandy2020towards,
  title={Towards maximizing the representation gap between in-domain \& out-of-distribution examples},
  author={Nandy, Jay and Hsu, Wynne and Lee, Mong Li},
  journal={Advances in neural information processing systems},
  volume={33},
  pages={9239--9250},
  year={2020}
}

@inproceedings{zhang2022implicit,
  title={Implicit sample extension for unsupervised person re-identification},
  author={Zhang, Xinyu and Li, Dongdong and Wang, Zhigang and Wang, Jian and Ding, Errui and Shi, Javen Qinfeng and Zhang, Zhaoxiang and Wang, Jingdong},
  booktitle={2022 IEEE/CVF Conference on Computer Vision and Pattern Recognition (CVPR)},
  pages={7359--7368},
  year={2022},
  organization={IEEE}
}

@inproceedings{litrico2023guiding,
  title={Guiding pseudo-labels with uncertainty estimation for source-free unsupervised domain adaptation},
  author={Litrico, Mattia and Del Bue, Alessio and Morerio, Pietro},
  booktitle={2023 IEEE/CVF Conference on Computer Vision and Pattern Recognition (CVPR)},
  pages={7640--7650},
  year={2023},
  organization={IEEE}
}

@inproceedings{dou2022reliability,
  title={Reliability-aware prediction via uncertainty learning for person image retrieval},
  author={Dou, Zhaopeng and Wang, Zhongdao and Chen, Weihua and Li, Yali and Wang, Shengjin},
  booktitle={European Conference on Computer Vision},
  pages={588--605},
  year={2022},
  organization={Springer}
}

@article{zhang2024vl,
  title={Vl-uncertainty: Detecting hallucination in large vision-language model via uncertainty estimation},
  author={Zhang, Ruiyang and Zhang, Hu and Zheng, Zhedong},
  journal={arXiv preprint arXiv:2411.11919},
  year={2024}
}

@inproceedings{lee2020gradients,
  title={Gradients as a measure of uncertainty in neural networks},
  author={Lee, Jinsol and AlRegib, Ghassan},
  booktitle={2020 IEEE International Conference on Image Processing (ICIP)},
  pages={2416--2420},
  year={2020},
  organization={IEEE}
}

@article{zhang2021uncertainty,
  title={Uncertainty-aware blind image quality assessment in the laboratory and wild},
  author={Zhang, Weixia and Ma, Kede and Zhai, Guangtao and Yang, Xiaokang},
  journal={IEEE Transactions on Image Processing},
  volume={30},
  pages={3474--3486},
  year={2021},
  publisher={IEEE}
}

@article{zheng2021rectifying,
  title={Rectifying pseudo label learning via uncertainty estimation for domain adaptive semantic segmentation},
  author={Zheng, Zhedong and Yang, Yi},
  journal={International Journal of Computer Vision},
  volume={129},
  number={4},
  pages={1106--1120},
  year={2021},
  publisher={Springer}
}

@article{her2023uncertainty,
  title={Uncertainty-aware gaze tracking for assisted living environments},
  author={Her, Paris and Manderle, Logan and Dias, Philipe A and Medeiros, Henry and Odone, Francesca},
  journal={IEEE Transactions on Image Processing},
  volume={32},
  pages={2335--2347},
  year={2023},
  publisher={IEEE}
}

@inproceedings{sabour2023robustnerf,
  title={Robustnerf: Ignoring distractors with robust losses},
  author={Sabour, Sara and Vora, Suhani and Duckworth, Daniel and Krasin, Ivan and Fleet, David J and Tagliasacchi, Andrea},
  booktitle={2023 IEEE/CVF Conference on Computer Vision and Pattern Recognition (CVPR)},
  pages={20626--20636},
  year={2023},
  organization={IEEE}
}

@inproceedings{
lipman2023flow,
title={Flow Matching for Generative Modeling},
author={Yaron Lipman and Ricky T. Q. Chen and Heli Ben-Hamu and Maximilian Nickel and Matthew Le},
booktitle={The Eleventh International Conference on Learning Representations },
year={2023},
url={https://openreview.net/forum?id=PqvMRDCJT9t}
}

@inproceedings{zhang2018unreasonable,
  title={The unreasonable effectiveness of deep features as a perceptual metric},
  author={Zhang, Richard and Isola, Phillip and Efros, Alexei A and Shechtman, Eli and Wang, Oliver},
  booktitle={2018 IEEE/CVF conference on computer vision and pattern recognition},
  pages={586--595},
  year={2018},
  organization={IEEE}
}
\bibliographystyle{conference}
\appendix

\clearpage

\newcommand{\apptocline}[3][0em]{%
\par\noindent\hspace*{#1}%
\hyperref[#2]{\makebox[2.2em][l]{\ref*{#2}}#3}%
\nobreak\dotfill\ \makebox[1.8em][r]{\pageref{#2}}\par
\vspace{2pt}%
}
\vspace*{-4em}
\begin{center}{\large\scshape Contents}\end{center}
\vspace{-0.4em}\hrule\vspace{0.6em}
{\small
\apptocline{sec:a}{Ablation on the Number of Perturbed Paths}
\apptocline{sec:b}{Ablation on Uncertainty Weighting Strength}
\apptocline{sec:c}{Ablation on Semantic Alignment Strength}
\apptocline{sec:d}{Ablation on Semantic Alignment Depth}
\apptocline{sec:e}{Additional Visual Comparison Results}
}
\vspace{0.4em}\hrule
\section{Ablation on the Number of Perturbed Paths}
\label{sec:a}
In our Dual-Path Target Construction, the consistency target $\hat{x}_0$ and the associated uncertainty proxy $U$ are computed from $K=2$ independently perturbed paths. To investigate the effect of the number of paths, we generalize the target construction to $K$ paths, where $\hat{x}_0 = \frac{1}{K}\sum_{i=1}^{K} f_\phi(\cdot)$, and evaluate different choices of $K\in\{1,2,3,4\}$. This experiment examines the trade-off between the stability of the averaged consistency target and the additional computational cost introduced by multiple paths. The corresponding VBench 2.0 scores are reported in Table~\ref{tab:s1}. The results show that using two paths provides an effective balance between target stability and computational overhead, motivating our default choice of $K=2$.

\section{Ablation on Uncertainty Weighting Strength}
\label{sec:b}
To investigate the effect of the uncertainty weighting strength $\lambda$, we evaluate six values, $\lambda\in\{0,0.1,0.5,1,5,10\}$, while keeping all other settings fixed. When $\lambda=0$, the weighting term reduces to $e^{-\lambda\tilde{U}}=1$, recovering the standard consistency distillation objective without uncertainty-aware reweighting. As shown in Table~\ref{tab:uw}, introducing uncertainty-aware weighting substantially improves performance, with the mean VBench 2.0 score increasing from 0.462 at $\lambda=0$ to 0.556 at $\lambda=1$. Larger values further improve the overall score, with $\lambda=5$ achieving the highest mean of 0.568, while $\lambda=10$ remains competitive at 0.562. These results indicate that the proposed uncertainty-aware supervision is beneficial over a broad range of weighting strengths. We use $\lambda=1$ as the default setting in our main experiments. This value is fixed a priori rather than selected on the benchmark: although $\lambda=5$ attains a higher mean, choosing the weighting strength on the same benchmark that is used for comparison would inflate the reported score, so Table~\ref{tab:uw} is provided to characterize the sensitivity rather than to select the operating point. We also note that, because $\tilde{U}$ is normalized by its per-video mean, a location with average uncertainty receives weight $e^{-\lambda}$, so $\lambda$ scales the overall magnitude of the distillation term relative to the adversarial term in the student objective in addition to controlling the relative reweighting.

\section{Ablation on Semantic Alignment Strength}
\label{sec:c}
The semantic alignment loss $\mathcal{L}_{\text{align}}$ anchors the discriminator's intermediate features to frozen vision embeddings, encouraging semantic consistency during few-step generation. We investigate the sensitivity of our framework to the alignment weight $\lambda_{\text{align}} \in \{0.1, 0.5, 1, 5, 10\}$. The corresponding VBench 2.0 scores are reported in Table~\ref{tab:s2}. We use $\lambda_{\text{align}}=1$ in all main experiments, which provides a favorable balance among the evaluated configurations.

\section{Ablation on Semantic Alignment Depth}
\label{sec:d}
Our semantic alignment loss $\mathcal{L}_{\text{align}}$ computes the cosine similarity between frozen vision embeddings and the discriminator's projected features. The projection heads take intermediate features extracted from selected layers of the Wan teacher model. To investigate the effect of feature extraction depth, we evaluate four configurations: the last layer only, the last 5 layers, the last 10 layers, and all layers. The corresponding VBench 2.0 scores are reported in Table~\ref{tab:s3}. Based on these results, we use features from all layers for semantic alignment in the main experiments.

\begin{table}[H]
  \centering
  \caption{Ablation on the number of perturbed paths.}
  \resizebox{0.90\linewidth}{!}{
    \small
    \begin{tabular}{c|ccccc|c}
    \toprule
    K     & Human Fidelity & Creativity & Controllability & Commonsense & Physics & Mean \\
    \midrule
    1     & 0.840  & 0.584  & 0.197  & 0.487  & 0.456  & 0.513  \\
   2     & \textbf{0.861} & 0.642  & \textbf{0.203} & 0.556  & 0.516  & \textbf{0.556} \\
    3     & 0.860  & 0.645  & 0.192  & 0.545  & \textbf{0.527} & 0.554  \\
    4     & 0.826  & \textbf{0.674} & 0.179  & \textbf{0.582} & 0.519  & \textbf{0.556} \\
    \bottomrule
    \end{tabular}%
    }
  \label{tab:s1}%
\end{table}%

\begin{table}[H]
  \centering
  \caption{Ablation on uncertainty weighting strength.}
  \resizebox{0.90\linewidth}{!}{
    \small
    \begin{tabular}{c|ccccc|c}
    \toprule
    $\lambda$     & Human Fidelity & Creativity & Controllability & Commonsense & Physics & Mean \\
    \midrule
    0     & 0.597  & 0.632  & 0.108  & 0.551  & 0.422  & 0.462  \\
    0.1   & 0.747  & 0.637  & 0.160  & \textbf{0.574} & 0.502  & 0.524  \\
    0.5   & 0.794  & 0.630  & 0.154  & 0.571  & 0.484  & 0.526  \\
    1     & 0.861  & 0.642  & \textbf{0.203} & 0.556  & 0.516  & 0.556  \\
    5     & \textbf{0.862} & 0.692  & 0.195  & \textbf{0.574} & 0.518  & \textbf{0.568} \\
    10    & 0.824  & \textbf{0.711} & 0.180  & 0.568  & \textbf{0.530} & 0.562  \\
    \bottomrule
    \end{tabular}%
    }
  \label{tab:uw}%
\end{table}%

\begin{table}[H]
  \centering
  \caption{Ablation on semantic alignment strength.}
  \resizebox{0.90\linewidth}{!}{
    \small
    \begin{tabular}{c|ccccc|c}
    \toprule
    $\lambda_{\text{align}}$ & Human Fidelity & Creativity & Controllability & Commonsense & Physics & Mean \\
    \midrule
    0.1   & 0.807  & 0.641  & 0.164  & 0.541  & 0.477  & 0.526  \\
    0.5   & 0.842  & 0.622  & 0.166  & 0.552  & 0.495  & 0.535  \\
    \textbf{1} & \textbf{0.861} & \textbf{0.642} & \textbf{0.203} & \textbf{0.556} & \textbf{0.516} & \textbf{0.556} \\
    5     & 0.838  & 0.640  & 0.189  & 0.528  & 0.493  & 0.538  \\
    10    & 0.812  & 0.611  & 0.158  & 0.553  & 0.488  & 0.524  \\
    \bottomrule
    \end{tabular}%
    }
  \label{tab:s2}%
\end{table}%

\begin{table}[H]
  \centering
  
  \caption{Ablation on semantic alignment depth.}
  \resizebox{0.90\linewidth}{!}{
    \small
    \begin{tabular}{c|ccccc|c}
    \toprule
    Depth & Human Fidelity & Creativity & Controllability & Commonsense & Physics & Mean \\
    \midrule
    1     & 0.860  & 0.627  & 0.166  & 0.555  & 0.490  & 0.540  \\
    5     & 0.853  & \textbf{0.647} & 0.185  & 0.548  & 0.514  & 0.549  \\
    10    & 0.844  & 0.613  & 0.190  & \textbf{0.591} & 0.506  & 0.549  \\
    all   & \textbf{0.861} & 0.642  & \textbf{0.203} & 0.556  & \textbf{0.516} & \textbf{0.556} \\
    \bottomrule
    \end{tabular}%
    }
  \label{tab:s3}%
\end{table}%

\section{Additional Visual Comparison Results}
\label{sec:e}

\begin{figure}[t]
 \centering
 \includegraphics[width=\linewidth]{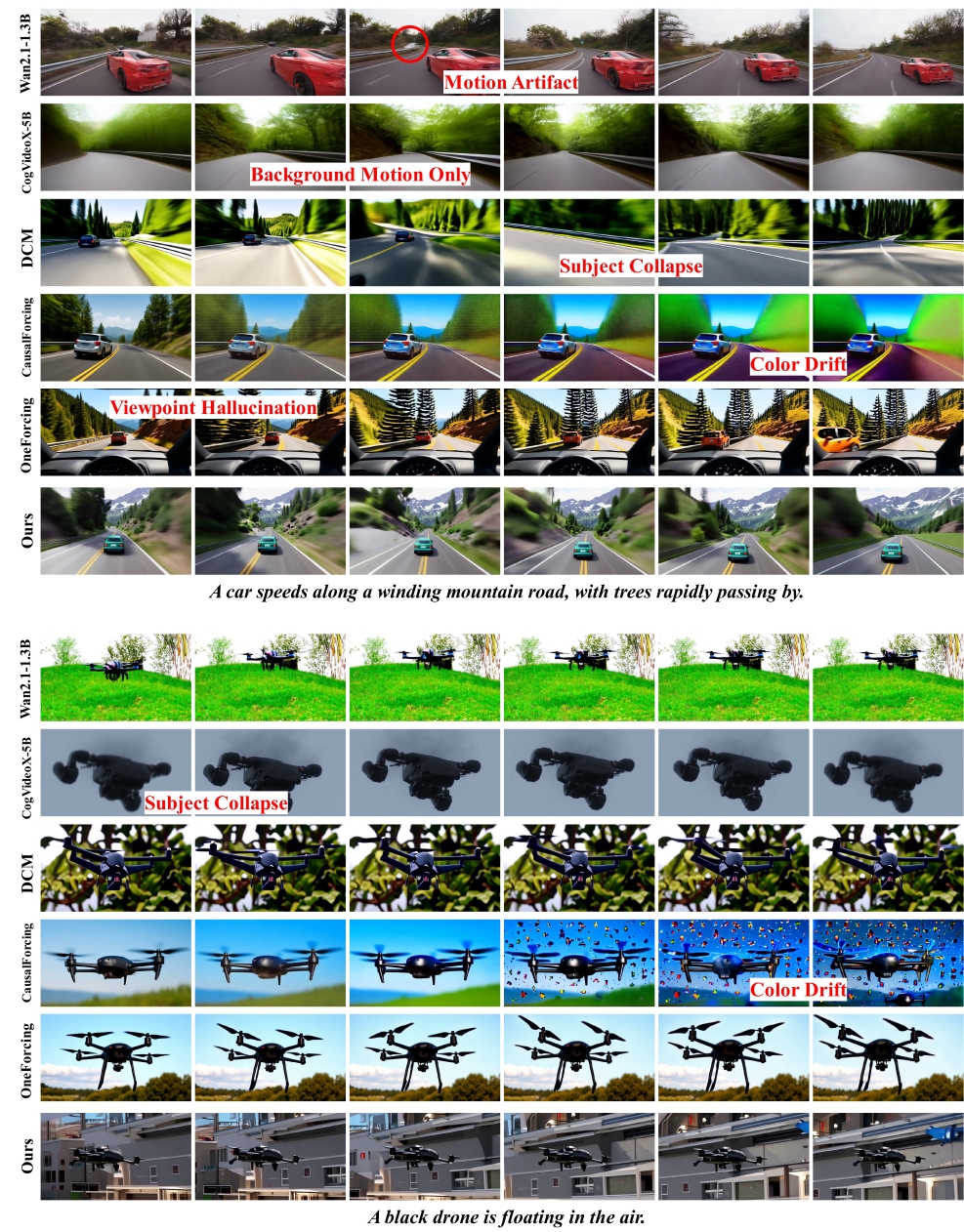}
 \vspace{-.2in}
 \caption{Additional qualitative comparison with competitive methods.}
 \label{fig:base2}
\end{figure}

\begin{figure}[t]
 \centering
 \includegraphics[width=\linewidth]{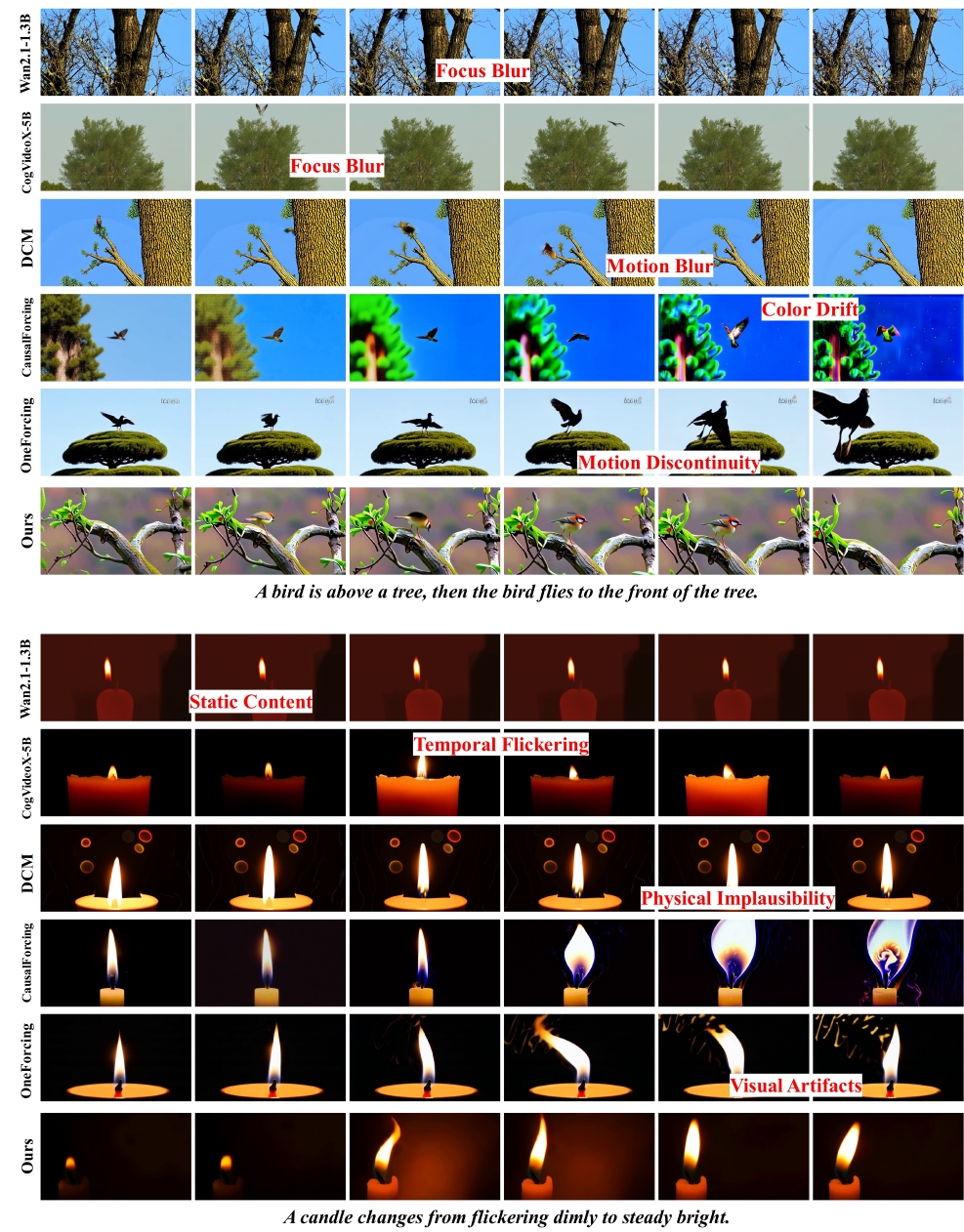}
 \vspace{-.2in}
 \caption{Additional qualitative comparison with competitive methods.}
 \label{fig:base3}
\end{figure}

\begin{figure}[t]
 \centering
 \includegraphics[width=\linewidth]{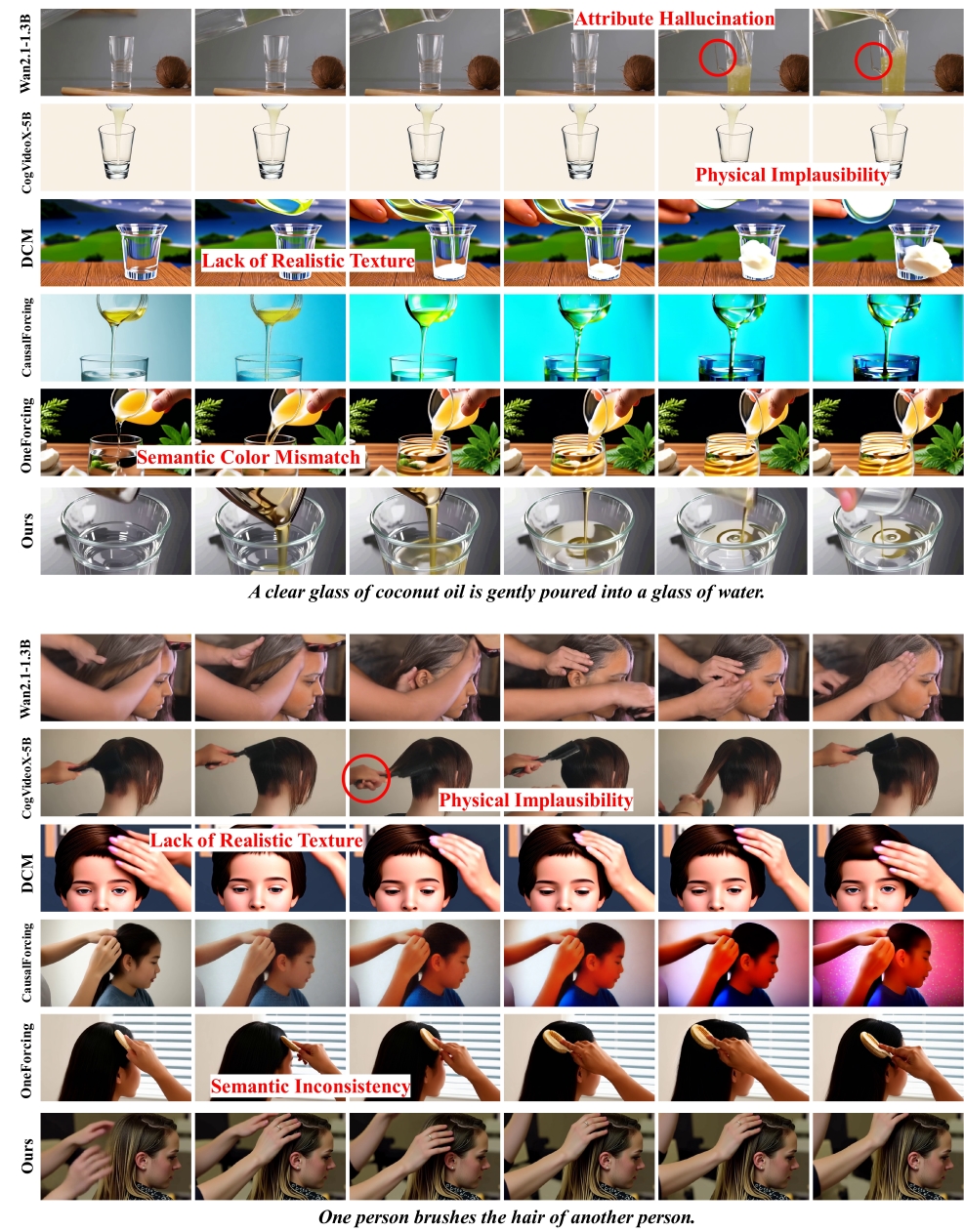}
 \vspace{-.2in}
 \caption{Additional qualitative comparison with competitive methods.}
 \label{fig:base4}
\end{figure}

More visual comparison results are presented in Figures~\ref{fig:base2}, \ref{fig:base3}, \ref{fig:base4}. We observe that our method consistently generates videos that align well with the given prompts while preserving temporal consistency, coherent motion, and stable subject appearance. These results further demonstrate the effectiveness of our uncertainty-aware supervision in improving video quality and maintaining spatiotemporal consistency under few-step generation.
More video samples are available on our anonymous project website: \url{https://uacd.github.io/UACD/}.

\end{document}